%% file: naaclhlt2019.tex
\documentclass[11pt,a4paper]{article}
\usepackage[hyperref]{naaclhlt2019}
\usepackage{times}
\usepackage{latexsym}

\usepackage{amsmath,amssymb,dsfont}
\usepackage{enumitem}
\usepackage{graphicx}
\usepackage[normalem]{ulem}
\usepackage{url}

\usepackage[linesnumbered,ruled,vlined,algo2e]{algorithm2e}

\usepackage{microtype}

\usepackage{multicol}
\usepackage{multirow}
\usepackage{array}
\usepackage{tabu,colortbl}
\usepackage{hhline}
\usepackage{subcaption}

\usepackage{xcolor}
 
\definecolor{our_blue}{rgb}{0.21747533000128158, 0.5305292836088684, 0.7548225041650647}
\definecolor{our_red}{rgb}{0.7364705882352941, 0.08, 0.10117647058823528}

\newcommand{\tr}[1][3pt]{\mathrel{%
   \hbox{\rule[\dimexpr\fontdimen22\textfont2-.2pt\relax]{#1}{.4pt}}%
   \mkern-4mu\hbox{\usefont{U}{lasy}{m}{n}\symbol{41}}}}

\aclfinalcopy % Uncomment this line for the final submission
\title{Competence-based Curriculum Learning for \\ Neural Machine Translation}

\author{Emmanouil Antonios Platanios$^\dagger$, Otilia Stretcu$^\dagger$, Graham Neubig$^\ddag$, Barnabas Poczos$^\dagger$, Tom M. Mitchell$^\dagger$ \\
        $^\dagger$Machine Learning Department, $^\ddag$Language Technologies Institute \\
        Carnegie Mellon University \\
        \texttt{\{e.a.platanios,ostretcu,gneubig,bpoczos,tom.mitchell\}@cs.cmu.edu}}

\date{}

\begin{document}
\maketitle
\begin{abstract}

Current state-of-the-art NMT systems use large neural networks that are not only slow to train, but also often require many heuristics and optimization tricks, such as specialized learning rate schedules and large batch sizes.
This is undesirable as it requires extensive hyperparameter tuning.
In this paper, we propose a {\em curriculum learning framework for NMT} that reduces training time, reduces the need for specialized heuristics or large batch sizes, and results in overall better performance.
Our framework consists of a principled way of deciding which training samples are shown to the model at different times during training, based on the estimated {\em difficulty} of a sample and the current {\em competence} of the model.
Filtering training samples in this manner prevents the model from getting stuck in bad local optima, making it converge faster and reach a better solution than the common approach of uniformly sampling training examples.
Furthermore, the proposed method can be easily applied to existing NMT models by simply modifying their input data pipelines.
We show that our framework can help improve the training time and the performance of both recurrent neural network models and Transformers, achieving up to a 70\% decrease in training time, while at the same time obtaining accuracy improvements of up to 2.2 BLEU.

\end{abstract}

\input{introduction}

\input{background}
\input{proposed_method}
\input{experiments}
\input{related_work}
\input{conclusion}

\section*{Acknowledgments}

We would like to thank Maruan Al-Shedivat and Dan Schwartz for the useful feedback they provided in early versions of this paper.
This research was supported in part by AFOSR under grant FA95501710218.

\bibliography{naaclhlt2019}
\bibliographystyle{acl_natbib}

\end{document}

%% file: introduction.tex
\section{Introduction}
\label{sec:intro}

Neural Machine Translation (NMT; \newcite{kalchbrenner2013recurrent,Bahdanau:2014}) now represents the state-of-the-art adapted in most machine translation systems \cite{Wu:2016,Crego:2016,bojar2017findings}, largely due to its ability to benefit from end-to-end training on massive amounts of data.
In particular, recently-introduced self-attentional Transformer architectures \cite{Vaswani:2017} are rapidly becoming the de-facto standard in NMT, having demonstrated both superior performance and training speed compared to previous architectures using recurrent neural networks (RNNs; \cite{kalchbrenner2013recurrent,sutskever2014sequence}).
However, large scale NMT systems are often hard to train, requiring complicated heuristics which can be both time-consuming and expensive to tune.
This is especially true for Transformers which, when carefully tuned, have been shown to consistently outperform RNNs \citep{Popel:2018}, but on the other hand, also rely on a number of heuristics such as specialized learning rates and large-batch training.

\begin{figure}[t]
	\centering
	\includegraphics[width=\columnwidth]{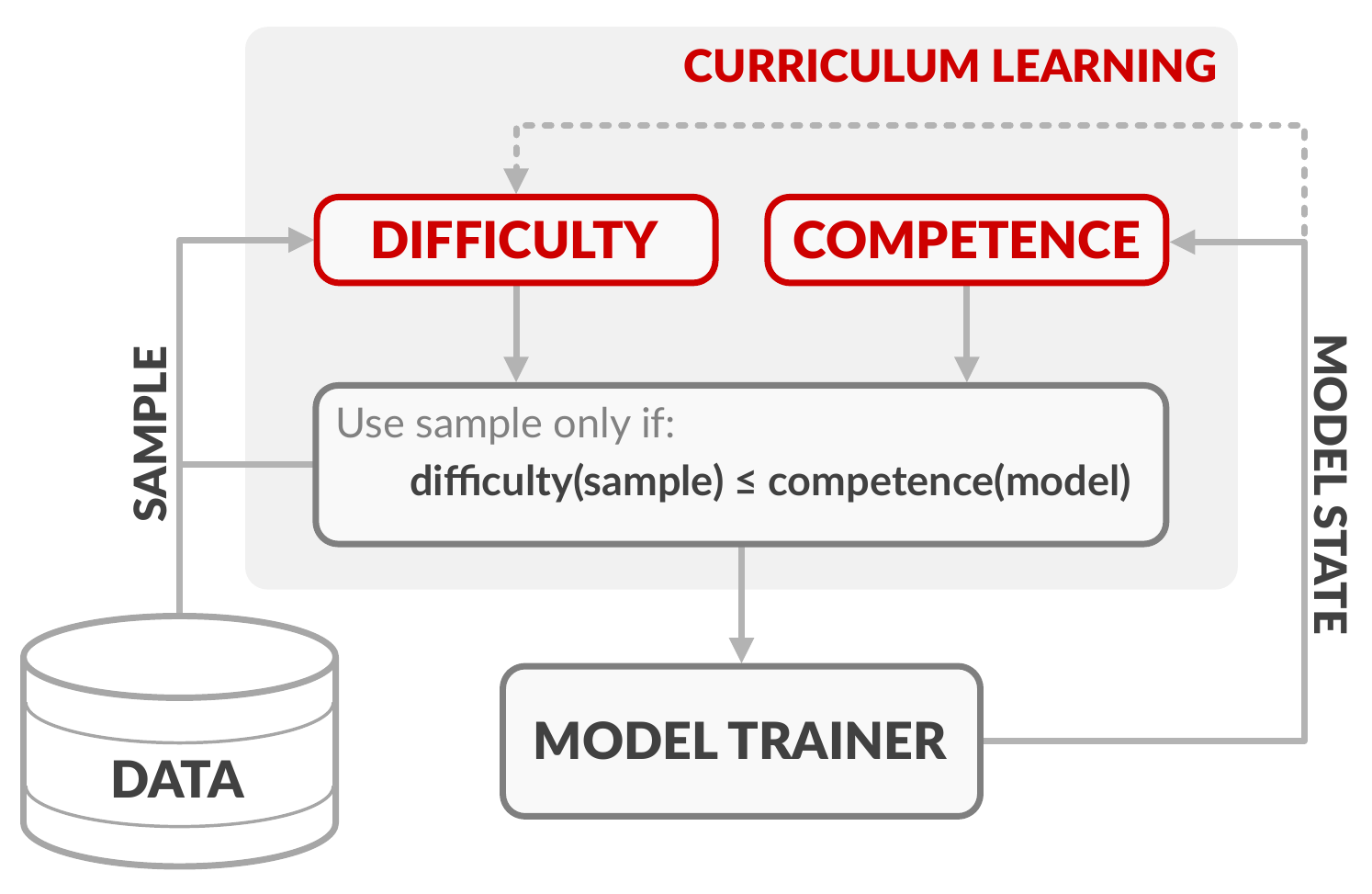}
	\caption{Overview of the proposed curriculum learning framework. During training, {\em difficulty} of each training sample is estimated and a decision whether to use it is made based on the current {\em competence} of the model.}
	\label{fig:overview}
%	\vspace{-0.8em}
\end{figure}

In this paper, we attempt to tackle this problem by proposing a {\em curriculum learning framework} for training NMT systems that reduces training time, reduces the need for specialized heuristics or large batch sizes, and results in overall better performance.
It allows us to train both RNNs and, perhaps more importantly, Transformers, with relative ease.
Our proposed method is based on the idea of teaching algorithms in a similar manner as humans, from easy concepts to more difficult ones.
This idea can be traced back to the work of \citet{elman1993learning} and \citet{krueger2009flexible}.
The main motivation is that training algorithms can perform better if training data is presented in a specific order, starting from {\em easy} examples and moving on to more {\em difficult} ones, as the learner becomes more {\em competent}.
In the case of machine learning, it can also be thought of as a means to avoid getting stuck in bad local optima early on in training.
An overview of the proposed framework is shown in Figure \ref{fig:overview}.

Notably, we are not the first to examine curriculum learning for NMT, although other related works have met with mixed success.
\citet{Kocmi:2017} explore impact of several curriculum heuristics on training a translation system for a single epoch, presenting the training examples in an easy-to-hard order based on sentence length and vocabulary frequency.
However, their strategy introduces all training samples during the first epoch, and how this affects learning in following epochs is not clear, with official evaluation results \citep{Bojar:2017} indicating that final performance may indeed be hurt with this strategy.
Contemporaneously to our work, \citet{Zhang:2018} further propose to split the training samples into a predefined number of bins (5, in their case), based on various difficulty metrics.
A manually designed curriculum schedule then specifies the bins from which the model samples training examples.
Experiments demonstrate that benefits of curriculum learning are highly sensitive to several hyperparameters (e.g., learning rate, number of iterations spent in each phase, etc.), and largely provide benefits in convergence speed as opposed to final model accuracy.

In contrast to these previous approaches, we define a continuous curriculum learning method (instead of a discretized regime) with only one tunable hyperparameter (the duration of curriculum learning).
Furthermore, as opposed to previous work which only focuses on RNNs, we also experiment with Transformers, which are notoriously hard to train \citep{Popel:2018}.
Finally, unlike any of the work described above, we show that our curriculum approach helps not only in terms of convergence speed, but also in terms of the learned model performance.
In summary, our method has the following desirable features:
\begin{enumerate}[noitemsep,leftmargin=13pt,topsep=2pt]
	\item \textbf{Abstract:} It is a novel, generic, and extensible formulation of curriculum learning. A number of previous heuristic-based approaches, such as that of \citet{Kocmi:2017}, can be formulated as special cases of our framework.
	\item \textbf{Simple:} It can be applied to existing NMT systems with only a small modification to their training data pipelines.
	\item \textbf{Automatic:} It does not require any tuning other than picking the value of a single parameter, which is the length of the curriculum (i.e., for how many steps to use curriculum learning, before easing into normal training).
	\item \textbf{Efficient:} It reduces training time by up to 70\%, whereas contemporaneous work of \citet{Zhang:2018} reports reductions of up to 46\%.
	\item \textbf{Improved Performance:} It improves the performance of the learned models by up to 2.2 BLEU points, where the best setting reported by \citet{Zhang:2018} achieves gains of up 1.55 BLEU after careful tuning.
\end{enumerate}
In the next section, we introduce our proposed curriculum learning framework.

%% file: proposed_method.tex
\section{Proposed Method}
\label{sec:proposed_method}

We propose {\em competence-based curriculum learning}, a training framework based on the idea that training algorithms can perform better if training data is presented in a way that picks examples appropriate for the model's current {\em competence}.
More specifically, we define the following two concepts that are central to our framework:

\paragraph{Difficulty:} A value that represents the difficulty of a training sample and that may depend on the current state of the learner. For example, sentence length is an intuitive difficulty metric for natural language processing tasks. The only constraint is that difficulty scores are comparable across different training samples (i.e., the training samples can be ranked according to their difficulty).

\begin{figure*}[t]
	\centering
	\includegraphics[width=\textwidth]{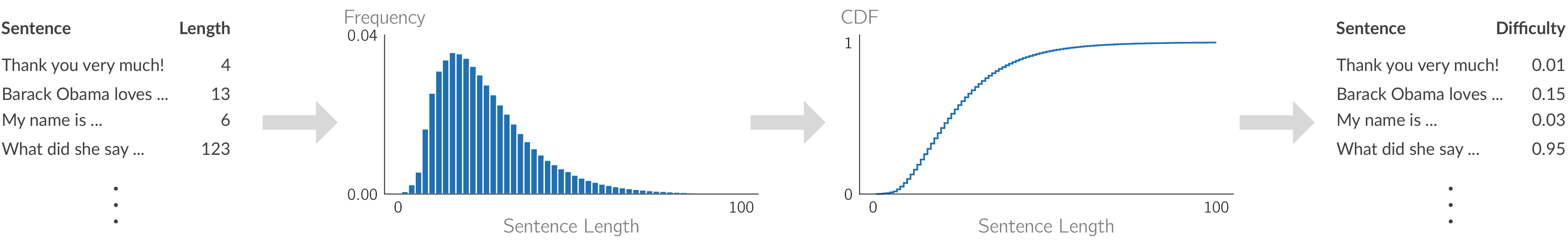}
	\caption{Example visualization of the preprocessing sequence used in the proposed algorithm. The histogram shown is that of sentence lengths from the WMT-16 \texttt{En}$\tr$\texttt{De} dataset used in our experiments. Here sentence lengths represent an example difficulty scoring function, $d$. ``CDF'' stands for the empirical ``cumulative density function'' obtained from the histogram on the left plot.}
	\label{fig:preprocessing_example}
%	\vspace{-0.8em}
\end{figure*}

\begin{figure}[t]
	\centering
	\includegraphics[width=\columnwidth]{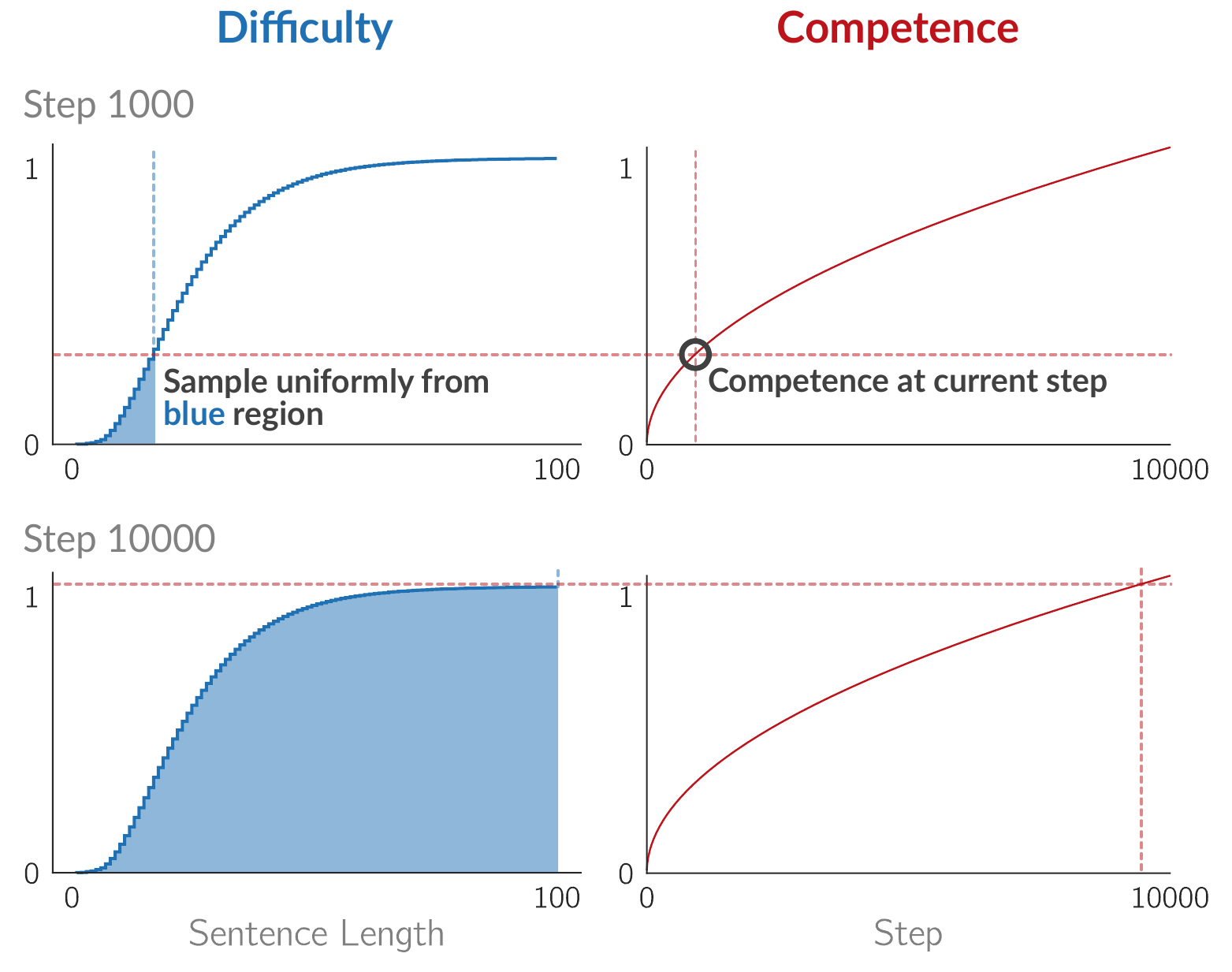}
	\caption{Example illustration of the training data ``filtering'' performed by our curriculum learning algorithm. At each training step: (i) the current competence of the model is computed, and (ii) a batch of training examples is sampled uniformly from all training examples whose difficulty is lower than that competence. In this example, we are using the sentence length difficulty heuristic shown in Equation~\ref{eq:sentence_length}, along with the square root competence model shown in Equation~\ref{eq:sqrt_competence}.}
	\label{fig:difficulty_competence_example}
%	\vspace{-0.8em}
\end{figure}

\paragraph{Competence:} A value between $0$ and $1$ that represents the progress of a learner during its training. It is defined as a function of the learner's state. More specifically, we define the competence, $c(t)$ at time $t$ (measured in terms of training steps), of a learner as the proportion of training data it is allowed to use at that time. The training examples are ranked according to their difficulty and the learner is only allowed to use the top $c(t)$ portion of them at time $t$.

%\vspace{0.5em}
\noindent Using these two concepts, we propose Algorithm~\ref{alg:ccl} (a high-level overview is shown in Figure \ref{fig:overview}, an example visualization of the first two steps is shown in Figure~\ref{fig:preprocessing_example}, and an example of the interaction between difficulty and competence is shown in Figure~\ref{fig:difficulty_competence_example}).

\begin{algorithm2e}[t]
\caption{Competence-based curriculum learning algorithm.}
\label{alg:ccl}
\footnotesize
\KwIn{Dataset, $\mathcal{D}=\{s_i\}_{i=1}^M$, consisting of $M$ samples, model trainer, $\mathcal{T}$, that takes as input batches of training data to use at each step, difficulty scoring function, $d$, and competence function, $c$.}
Compute the difficulty, $d(s_i)$, for each $s_i \in \mathcal{D}$. \\
Compute the cumulative density function (CDF) of the difficulty scores. This results in one difficulty CDF score per sample, $\bar{d}(s_i) \in [0,1]$. Illustrated in Figure~\ref{fig:preprocessing_example}. \\
    \For{training step $t=1,\hdots$}{
    	Compute the model competence, $c(t)$. \\
    	Sample a data batch, $B_t$, uniformly from all $s_i \in \mathcal{D}$, such that $\bar{d}(s_i) \leq c(t)$. Illustrated in Figure~\ref{fig:difficulty_competence_example}. \\
    	Invoke the trainer, $\mathcal{T}$, using $B_t$ as input.
    }
\KwOut{Trained model.}
\end{algorithm2e}

Note that, at each training step, we are not changing the relative probability of each training sample under the input data distribution, but we are rather constraining the domain of that distribution, based on the current competence of the learner.
Eventually, once the competence becomes $1$, the training process becomes equivalent to that without using a curriculum, with the main difference that the learner should now be more capable to learn from the more difficult examples.
Given the dependence of this algorithm on the specific choices of the difficulty scoring function, $d$, and the competence function, $c$, we now describe our instantiations for training NMT models.

\subsection{Difficulty Metrics}

There are many possible ways of defining the difficulty of translating a sentence.
We consider two heuristics inspired by what we, as humans, may consider difficult when translating, and by factors which can negatively impact the optimization algorithms used when training NMT models.
% \gn{This is just based on your intuition, right? Ideally when you say things like this it's good to have a citation.}
In the rest of this section we denote our training corpus as a collection of $M$ sentences, $\{s_i\}_{i=1}^M$, where each sentence is a sequence of words. $s_i = \{w_0^i,\hdots,w_{N_i}^i\}$.

\paragraph{Sentence Length:}
We argue that it is harder to translate longer sentences, as longer sentences require being able to translate their component parts, which often consist of short sentences.
Furthermore, longer sentences are intuitively harder to translate due to the propagation of errors made early on when generating the target language sentence.
% Also, when considering models such as the Transformer, sentence length can have a large impact on training due to the fact that the model attends over each part of the sentence at all stages of processing \gn{Is this empirically proven somewhere? Or just intuition? If the former then cite a paper or your experiments. If the latter, then you might want to explain more carefully. RNNs also use information from all parts of the sentence, so the difference may not be clear.}.
% This means that long sentences can result in high gradient variance, and thus make training unstable \gn{Really? Where's the evidence?}.
Therefore, a simple way to define the difficulty of a sentence $s_i = \{w_0^i,\hdots,w_{N_i}^i\}$ is as follows:
\begin{equation}
	d_{\textrm{length}}(s_i) \triangleq N_i.
	\label{eq:sentence_length}
\end{equation}
Note that we can compute this difficulty metric on either the source language sentence or the target language sentence.
We only consider the source sentence in this paper
\footnote{NMT models typically first pick up information about producing sentences of correct length.
It can be argued that presenting only short sentences first may lead to learning a strong bias for the sentence lengths.
In our experiments, we did not observe this to be an issue as the models kept improving and predicting sentences of correct length, throughout training.}.

\paragraph{Word Rarity:}
Another aspect of language that can affect the difficulty of translation is the frequency with which words appear.
For example, humans may find rare words hard to translate because we rarely ever see them and it may be hard to recall their meaning.
% \gn{This human analogy seems squishy, is it really that simple? Often for rare words it's as simple as looking them up in the dictionary, as they only have one translation. I would argue rather that it's harder to translate ambiguous or emotional words, even if they're common. It'd be good to have a citation.}.
The same can be true for NMT models where: (i) the statistical strength of the training examples containing rare words is low and thus the model needs to keep revisiting such words in order to learn robust representations for them, and (ii) the gradients of the rare word embeddings tend to have high variance; they are overestimates of the true gradients in the few occasions where they are non-zero, and underestimates otherwise.
%\gn{Really? Where's the evidence? Also, in the human example you're saying ``because we rarely ever see them'', but here you're talking about gradients, so there seems to be a disconnect.}.
This suggests that using word frequencies may be a helpful difficulty heuristic.
Given a corpus of sentences, $\{s_i\}_{i=1}^M$, we define relative word frequencies as:
\begin{equation}
	\hat{p}(w_j) \triangleq \frac{1}{N_{\textrm{total}}} \sum_{i=1}^{M} \sum_{k=1}^{N_i} \mathds{1}_{w_k^i = w_j},
\end{equation}
where $j = 1, \hdots, \textrm{\#\{unique words in corpus\}}$ and $\mathds{1}_{\textrm{condition}}$ is the indicator function which is equal to $1$ if its condition is satisfied and $0$ otherwise.
Next we need to decide how to aggregate the relative word frequencies of all words in a sentence to obtain a single difficulty score for that sentence.
Previous research has proposed various pooling operations, such as minimum, maximum, and average \citep{Zhang:2018}, but they show that they do not work well in practice.
We propose a different approach.
Ultimately, what might be most important is the overall likelihood of a sentence as that contains information about both word frequency and, implicitly, sentence length.
An approximation to this likelihood is the product of the unigram probabilities, which is related to previous work in the area of active learning \citep{settles2008activelearning}.
This product can be thought of as an approximate language model (assuming words are sampled independently) and also implicitly incorporates information about the sentence length that was proposed earlier (longer sentence scores are products over more terms in $[0, 1]$ and are thus likely to be smaller).
We thus propose the following difficulty heuristic:
\begin{equation}
	d_{\textrm{rarity}}(s_i) \triangleq -\sum_{k=1}^{N_i} \log \hat{p}(w_k^i),
\end{equation}
where we use logarithms of word probabilities to prevent numerical errors.
Note that negation is used because we define less likely (i.e., more rare) sentences as more difficult.

These are just two examples of difficulty metrics, and it is easy to conceive of other metrics such as the occurrence of homographs \cite{liu2018homographs} or context-sensitive words \cite{bawden2018discourse}, the examination of which we leave for future work.

\subsection{Competence Functions}

For this paper, we propose two simple functional forms for $c(t)$ and justify them with some intuition.
More sophisticated strategies that depend on the loss function, the loss gradient, or on the learner's performance on held-out data, are possible, but we do not consider them in this paper.

% \gn{These decay strategies are very similar to the ones used in \cite{bengio2015scheduled} (the figure was so reminiscent of the paper that I was surprised to not see a citation). I'd cite this.}
% I don't think this is super relevant. They only define linear, sigmoid, and exponential decay functions, similar to what's previously been used in many other areas, such as optimization, but the motivation is very different and the way they arrive to them is different (they basically just state they use these three specific functions). Only the linear function has a similar form, but that is just a linear function. Our exponential decay is different than anything they use and we arrive to it in a different way too.

\paragraph{Linear:} This is a simple way to define $c(t)$. 
Given an initial value $c_0 \triangleq c(0) \geq 0$ and a slope parameter $r$, we define:
\begin{equation}
	c(t) \triangleq \min\left( 1, tr + c_0 \right).
\end{equation}
In this case, new training examples are constantly being introduced during the training process, with a constant rate $r$ (as a proportion of the total number of available training examples). 
Note that we can also define $r = (1 - c_0) / T$, where $T$ denotes the time after which the learner is fully competent, which results in:
\begin{equation}
	c_{\textrm{linear}}(t) \triangleq \min\left( 1, t\frac{1 - c_0}{T} + c_0 \right).
	\label{eq:linear_competence}
\end{equation}

\begin{figure}[t]
	\centering
	\includegraphics[width=0.9\columnwidth]{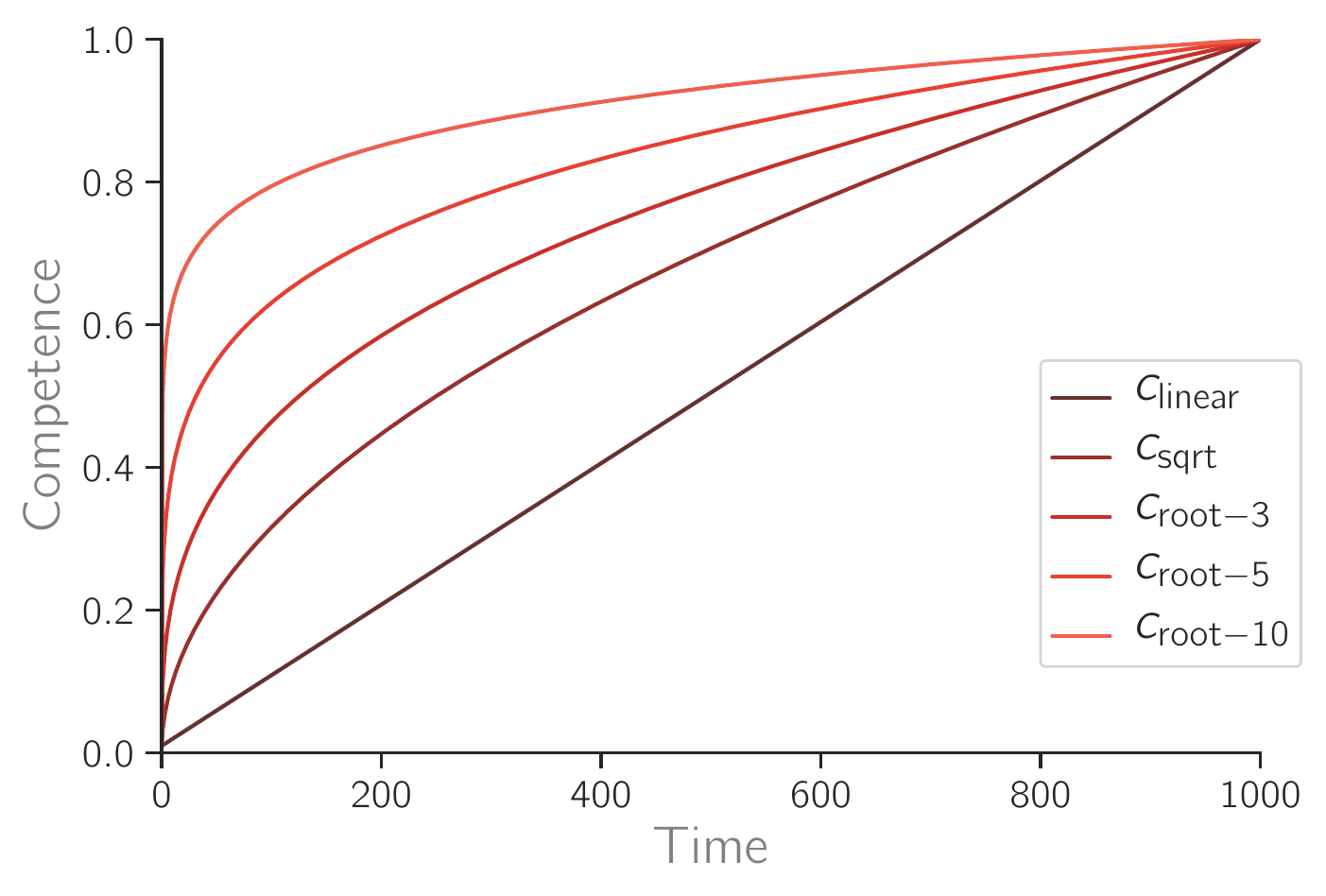}
%	\vspace{-0.8em}
	\caption{Plots of various competence functions with $c_0 = 0.01$ (initial competence value) and $T = 1,000$ (total duration of the curriculum learning phase).}
	\label{fig:competence}
%	\vspace{-0.8em}
\end{figure}

\paragraph{Root:} In the case of the linear form, the same number of new and more difficult, examples are added to the training set, at all times $t$. 
However, as the training data grows in size, it gets less likely that any single data example will be sampled in a training batch. 
Thus, given that the newly added examples are less likely to be sampled, we propose to reduce the number of new training examples per unit time as training progresses to give the learner sufficient time to assimilate their information content. 
More specifically, we define the rate in which new examples are added as inversely proportional to the current training data size:
\begin{equation}
	\frac{dc(t)}{dt} = \frac{P}{c(t)},	
\end{equation}
for some constant $P \geq 0$. Solving this simple differential equation, we obtain:
\begin{equation*}
	\int c(t) dc(t) = \int P dt \Rightarrow c(t) = \sqrt{2Pt + D},
\end{equation*}
for some constants $P$ and $D$. Then, we consider the following constraint: $c_0 \triangleq c(0) = \sqrt{D} \Rightarrow D = c_0^2$. 
Finally, we also have that $c(T) = 1 \Rightarrow P = (1-c_0^2) / 2T$, where $T$ denotes the time after which the learner is fully competent. 
This, along with the constraint that $c(t) \in [0, 1]$ for all $t \geq 0$, results in the following definition:
\begin{equation}
	c_{\textrm{sqrt}}(t) \triangleq \min\left( 1, \sqrt{t \frac{1 - c_0^2}{T} + c_0^2} \right).
	\label{eq:sqrt_competence}
\end{equation}
In our experiments, we refer to this specific formulation as the ``square root'' competence model.
If we want to make the curve {\em sharper}, meaning that even more time is spent per sample added later on in training, then we can consider the following more general form, for $p \geq 1$:
\begin{equation}
	c_{\textrm{root-}p}(t) \triangleq \min\left( 1, \sqrt[\uproot{3}p]{t \frac{1 - c_0^p}{T} + c_0^p} \right).
\end{equation}
We observed that best performance is obtained when $p=2$ and then, as we increase $p$, performance converges to that obtained when training without a curriculum.
Plots of the competence functions we presented are shown in Figure \ref{fig:competence}.

\subsection{Scalability}

Our method can be easily used in large-scale NMT systems.
This is because it mainly consists of a preprocessing step of the training data that computes the difficulty scores.
The implementation we are releasing with this paper computes these scores in an efficient manner by building a graph describing their dependencies, as well as whether they are sentence-level scores (e.g., sentence length), or corpus-level (e.g., CDF), and using that graph to optimize their execution.
Using only 8GB of memory, we can process up to 20k sentences per second when computing sentence rarity scores, and up to 150k sentences per second when computing sentence length scores.

%% file: experiments.tex
\section{Experiments}

For our experiments, we use three of the most commonly used datasets in NMT, that range from a small benchmark dataset to a large-scale dataset with millions of sentences.
Statistics about the datasets are shown in Table~\ref{tab:exp:dataset_sizes}.
We perform experiments using both RNNs and Transformers.
For the RNN experiments we use a bidirectional LSTM for the encoder, and an LSTM with the attention model of \citet{Bahdanau:2014} for the decoder.
The number of layers of the encoder and the decoder are equal.
We use a 2-layer encoder and a 2-layer decoder for all experiments on IWSLT datasets, and a 4-layer encoder and a 4-layer decoder for all experiments on the WMT dataset, due to the dataset's significantly larger size.
For the Transformer experiments we use the \textsc{Base} model proposed by \citet{Vaswani:2017}.
It consists of a 6-layer encoder and decoder, using 8 attention heads, and 2,048 units for the feed-forward layers.
The multi-head attention keys and values depth is set to the word embedding size.
The word embedding size is 512 for all experiments.
Furthermore, for the Transformer experiments on the two smaller datasets we do not use any learning rate schedule, and for the experiments on the largest dataset we use the default Transformer schedule.
A detailed discussion on learning rate schedules for Transformers is provided near the end of this section.
All of our experiments were conducted on a machine with a single Nvidia V100 GPU, and 24 GBs of system memory.

During training, we use a label smoothing factor of 0.1 \citep{Wu:2016} and the AMSGrad optimizer \cite{Reddi:2018} with its default parameters in TensorFlow, and a batch size of 5,120 tokens (due to GPU memory constraints).
During inference, we employ beam search with a beam size of 10 and the length normalization scheme of \citet{Wu:2016}.%
\footnote{We emphasize that we did not run experiments with other architectures or configurations, and thus our baseline architectures were not chosen because they were favorable to our method, but rather because they were frequently mentioned in existing literature.}

\begin{table}[t]
	\begin{center}
		\begin{tabu} to \columnwidth {|@{}>{}X[2r]||@{}>{}X[c]@{}|@{}>{}X[c]@{}|@{}>{}X[c]@{}|}
		\hhline{-||---}
		Dataset & \# Train & \# Dev & \# Test \\ \hhline{=::===}
		IWSLT-15 \texttt{En}$\tr$\texttt{Vi} & $133\textrm{k}$ & $\phantom{0}768$ & $1268$ \\ % dev2010, tst2013
		IWSLT-16 \texttt{Fr}$\tr$\texttt{En} & $224\textrm{k}$ & $1080$           & $1133$ \\ % tst2015, tst2016
		WMT-16 \texttt{En}$\tr$\texttt{De}   & $4.5\textrm{m}$ & $3003$           & $2999$ \\ % newstest2014 newstest2016
		\hhline{-||---}
		\end{tabu}
	\end{center}
%	\vspace{-0.5em}
	\caption{Number of parallel sentences in each dataset. ``k'' stands for ``thousand'' and ``m'' stands for ``million''.}
	\label{tab:exp:dataset_sizes}
%	\vspace{-1.0em}
\end{table}

\paragraph{Curriculum Hyperparameters.}

We set the initial competence $c_0$ to 0.01, in all experiments.
This means that all models start training using the 1\% easiest training examples.
The curriculum length $T$ is effectively the only hyperparameter that we need to set for our curriculum methods.
In each experiment, we set $T$ in the following manner: we train the baseline model without using any curriculum and we compute the number of training steps it takes to reach approximately 90\% of its final BLEU score.
We then set $T$ to this value.
This results in $T$ being set to 5,000 for the RNN experiments on the IWSLT datasets, and 20,000 for the corresponding Transformer experiments.
For WMT, we set $T$ to 20,000 and 50,000 for RNNs and Transformers, respectively.
Furthermore, we use the following notation and abbreviations when presenting our results:
\begin{itemize}[noitemsep,leftmargin=15pt,topsep=2pt,label=--]
	\item \uline{Plain:} Trained without using any curriculum.
	\item \uline{SL:} Curriculum with sentence length difficulty.
	\item \uline{SR:} Curriculum with sentence rarity difficulty.
	\item \uline{Linear}: Curriculum with the linear competence shown in Equation~\ref{eq:linear_competence}.
	\item \uline{Sqrt}: Curriculum with the square root competence shown in Equation~\ref{eq:sqrt_competence}.
\end{itemize}

\paragraph{Data Preprocessing.}

Our experiments are performed using the machine translation library released by \citet{Platanios:2018}.
We use the same data preprocessing approach the authors used in their experiments.
While training, we consider sentences up to length 200.
Similar to them, for the IWSLT-15 experiments we use a per-language vocabulary which contains the 20,000 most frequently occurring words, while ignoring words that appear less than 5 times in the whole corpus.
For the IWSLT-16 and WMT-16 experiments we use a byte-pair encoding (BPE) vocabulary \citep{Sennrich:2016a} trained using 32,000 merge operations, similar to the original Transformer paper by \citet{Vaswani:2017}.

\begin{table*}[t!]
	\begin{center}
		\begin{tabu} to \textwidth {|@{}>{}X[c]@{}|@{}>{}X[2c]@{}||@{}>{}X[2c]@{}|@{}>{}X[2c]@{}|@{}>{}X[2c]@{}|@{}>{}X[2c]@{}|@{}>{}X[2c]@{}||@{}>{}X[2c]@{}|@{}>{}X[2c]@{}|@{}>{}X[2c]@{}|@{}>{}X[2c]@{}|@{}>{}X[2c]@{}|@{}>{}X[2c]@{}|}

		\hhline{~~|-----||------}

		\multicolumn{2}{c|}{} & \multicolumn{5}{c||}{\textsc{RNN}} & \multicolumn{6}{c|}{\textsc{Transformer}} \\ 
		
		\hhline{~~-----||------}
		
		\multicolumn{2}{c|}{} & \multirow{2}{*}{\small Plain} & \multicolumn{2}{c|}{\small SL Curriculum} & \multicolumn{2}{c||}{\small SR Curriculum} & \multirow{2}{*}{\small Plain} & \multirow{2}{*}{\small Plain*} & \multicolumn{2}{c|}{\small SL Curriculum} & \multicolumn{2}{c|}{\small SR Curriculum} \\
		
		\hhline{~~~----||~~----}
		
		\multicolumn{2}{c|}{} & & $c_{\textrm{linear}}$ & $c_{\textrm{sqrt}}$ & $c_{\textrm{linear}}$ & $c_{\textrm{sqrt}}$ & & & $c_{\textrm{linear}}$ & $c_{\textrm{sqrt}}$ & $c_{\textrm{linear}}$ & $c_{\textrm{sqrt}}$ \\
		
		\hhline{--::=====::======}
		
		\multirow{3}{*}{\rotatebox[origin=c]{90}{\small BLEU}} & \texttt{En}$\tr$\texttt{Vi} & 26.27 & 26.57 & \textbf{27.23} & 26.72 & 26.87 & 28.06 & 29.77 & 29.14 & 29.57 & 29.03 & \textbf{29.81} \\
		& \texttt{Fr}$\tr$\texttt{En} & 31.15 & 31.88 & \textbf{31.92} & 31.39 & 31.57 & 34.05 & 34.88 & 34.98 & 35.47 & 35.30 & \textbf{35.83} \\
		& \texttt{En}$\tr$\texttt{De} & 26.53 & 26.55 & 26.54 & \textbf{26.62} & \textbf{26.62} & -- & 27.95 & 28.71 & 29.28 & 29.93 & \textbf{30.16} \\
		
		\hhline{==::=====::======}
		
		\multirow{3}{*}{\rotatebox[origin=c]{90}{\small Time}} & \texttt{En}$\tr$\texttt{Vi} & 1.00 & 0.64 & 0.61 & 0.71 & \textbf{0.57} & 1.00 & 1.00 & 0.44 & 0.33 & 0.35 & \textbf{0.31} \\
		& \texttt{Fr}$\tr$\texttt{En} & 1.00 & 1.00 & 0.93 & 1.10 & \textbf{0.73} & 1.00 & 1.00 & 0.49 & 0.44 & 0.42 & \textbf{0.39} \\
		& \texttt{En}$\tr$\texttt{De} & 1.00 & 0.86 & 0.89 & 1.00 & \textbf{0.83} & -- & 1.00 & 0.58 & \textbf{0.55} & \textbf{0.55} & \textbf{0.55} \\
		\hhline{--||-----||------}
		\end{tabu}
	\end{center}
%	\vspace{-0.8em}
	\caption{Summary of experimental results. For each method and dataset, we present the test set BLEU score of the best model based on validation set performance. We also show the relative time required to obtain the BLEU score of the best performing baseline model. For example, if an RNN gets to $26.27$ BLEU in $10,000$ steps and the SL curriculum gets to the same BLEU in $3,000$ steps, then the plain model gets a score of $1.0$ and the SL curriculum receives a score of $3,000 / 10,000 = 0.3$. ``Plain'' stands for the model trained without a curriculum and, for Transformers, ``Plain*'' stands for the model trained using the learning rate schedule shown in Equation \ref{eq:noam_schedule}.}
	\label{tab:results}
%	\vspace{-0.8em}
\end{table*}

\paragraph{Results.}

\begin{figure}[t!]
	\centering
    \includegraphics[width=\columnwidth]{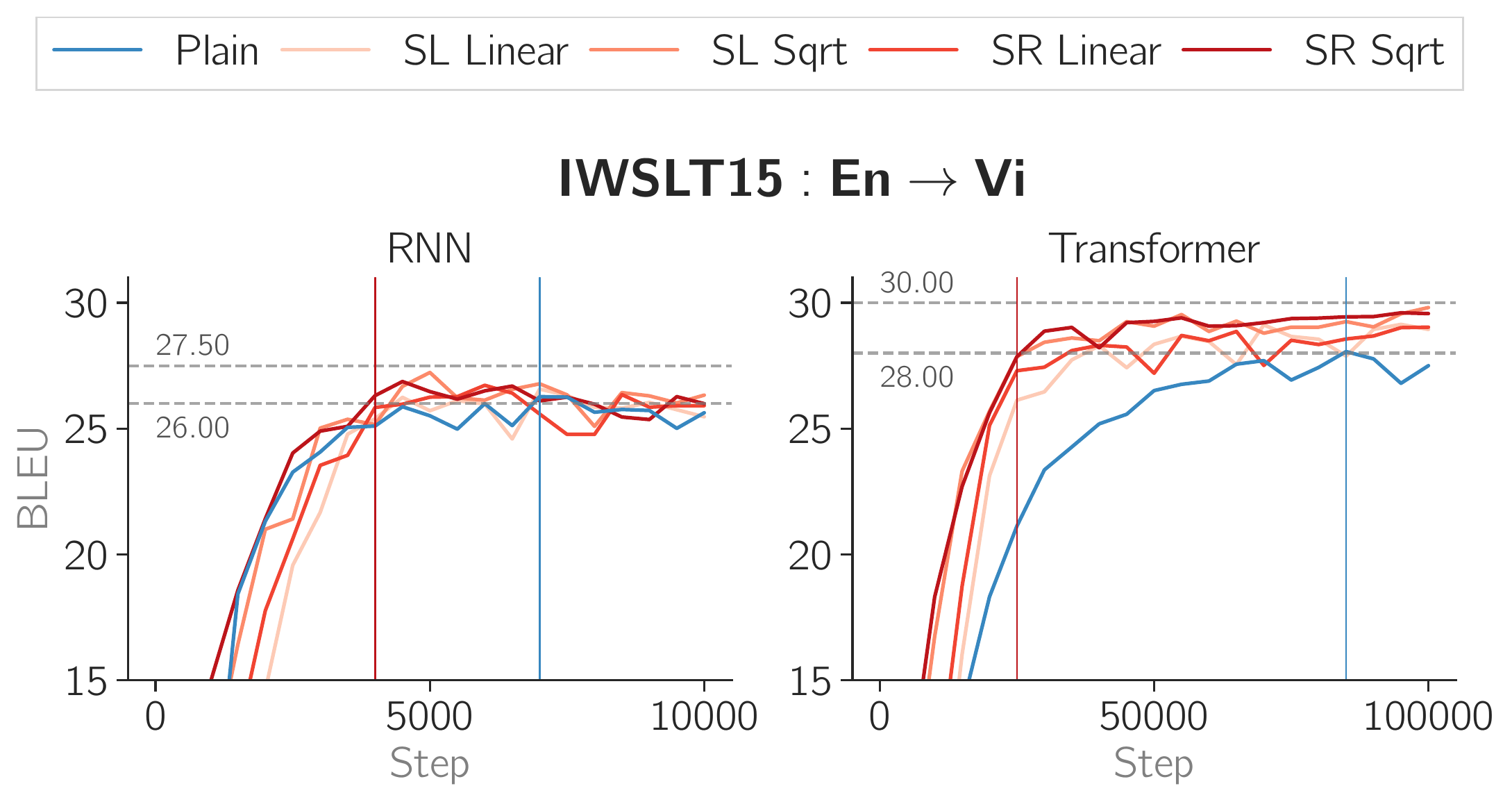}
    \includegraphics[width=\columnwidth]{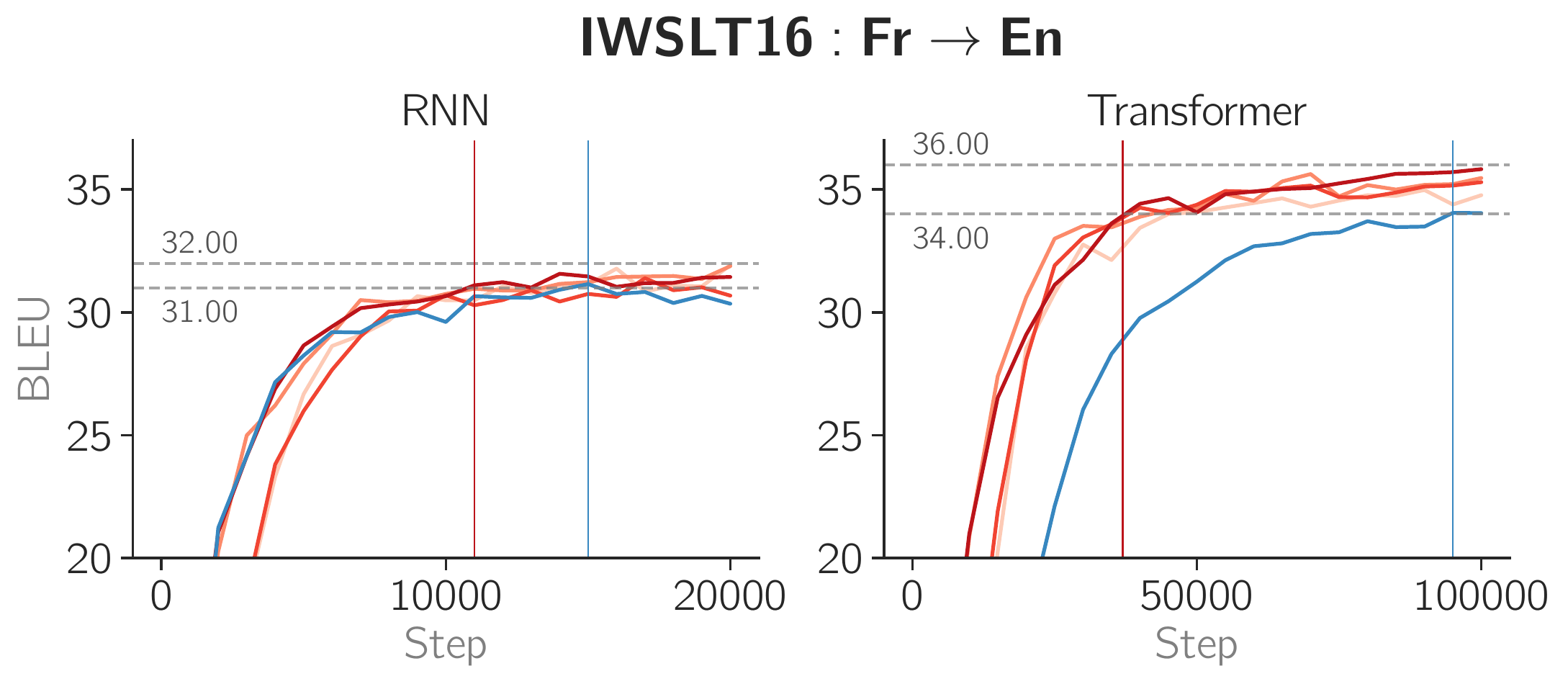}
    \includegraphics[width=\columnwidth]{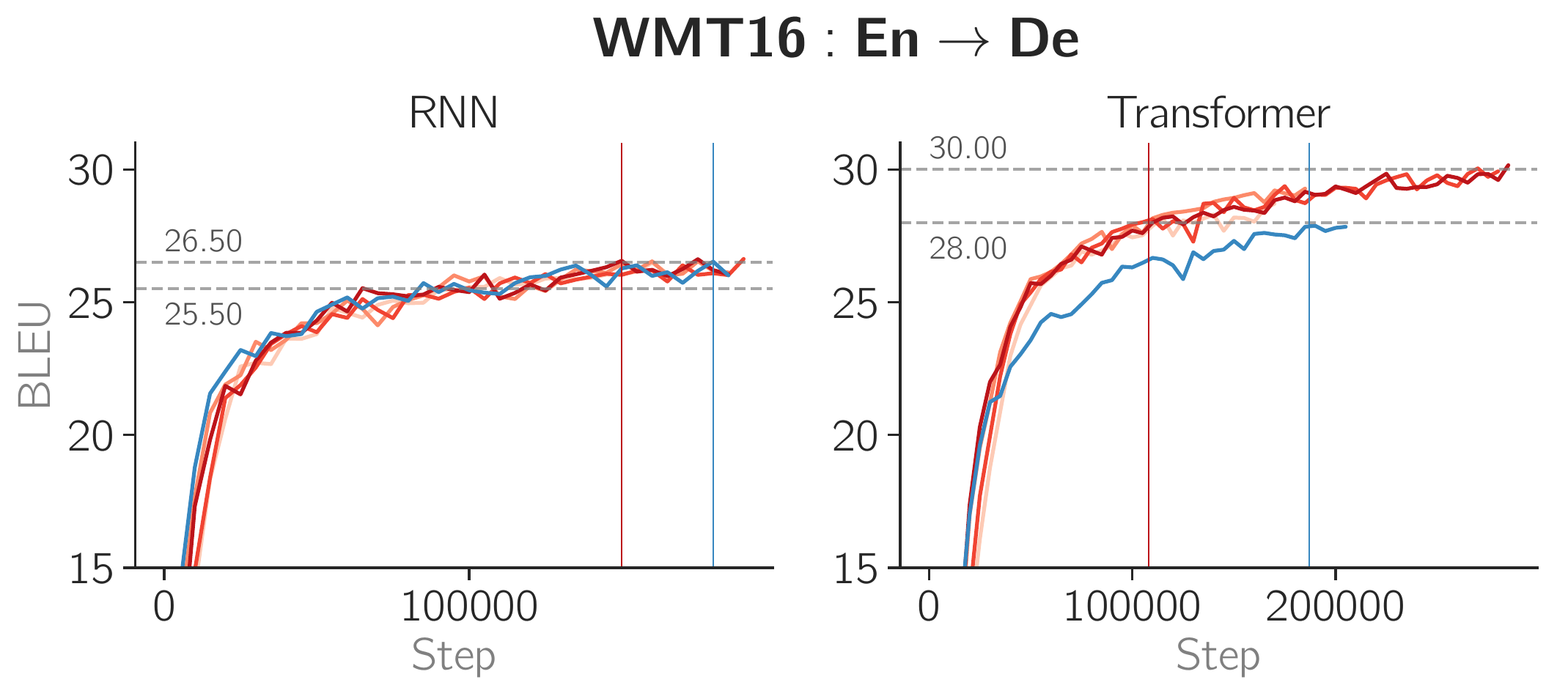}
    \caption{Plots illustrating the performance of various models on the test set, as training progresses. \textcolor{our_blue}{\textbf{Blue}} lines represent the baseline methods when no curriculum is used, and \textcolor{our_red}{\textbf{red}} lines represent the same models when different versions of our curriculum learning framework are used to train them. The vertical lines represent the step in which the models attain the BLEU score that the baseline models attain at convergence.}
    \label{fig:results}
%    \vspace{-0.7em}
\end{figure}

We present a summary of our results in Table~\ref{tab:results} and we also show complete learning curves for all methods in Figure~\ref{fig:results}.
The evaluation metrics we use are the test set BLEU score and the time it takes for the models using curriculum learning to obtain the BLEU score that the baseline models attain at convergence.
We observe that Transformers consistently benefit from our curriculum learning approach, achieving gains of up to 2 BLEU, and reductions in training time of up to 70\%. RNNs also benefit, but to a lesser extent.
This is consistent with our motivation for this paper, which stems from the observation that training RNNs is easier and more robust than training Transformers.
Furthermore, the square root competence model consistently outperforms the linear model, which fits well with our intuition and motivation for introducing it.
Regarding the difficulty heuristics, sentence length and sentence rarity both result in similar performance.

We also observe that, for the two small datasets, RNNs converge faster than Transformers in terms of both the number of training iterations and the overall training time.
This is contrary to other results in the machine translation community \citep[e.g.,][]{Vaswani:2017}, but could be explained by the fact that we are not using any learning rate schedule for training Transformers.
However, they never manage to outperform Transformers in terms of test BLEU score of the final model.
Furthermore, to the best of our knowledge, for IWSLT-15 we achieve state-of-the-art performance.
The highest previously reported result was $29.03$ BLEU \citep{Platanios:2018}, in a multi-lingual setting.
Using our curriculum learning approach we are able to achieve a BLEU score of $29.81$ for this dataset.

Overall, we have shown that {\em our curriculum learning approach consistently outperforms models trained without any curriculum}, in both limited data settings and large-scale settings.

\paragraph{Learning Rate Schedule.}

In all of our IWSLT experiments so far, we use the default AMSGrad learning rate of 0.001 and intentionally avoid using any learning rate schedules.
However, Transformers are not generally trained without a learning rate schedule, due to their instability.
Such schedules typically use a warm-up phase, which means that the learning rate starts at a very low value and keeps increasing until the end of the warm-up period, after which a decay rate is typically used.
In order to show that our curriculum learning approach can act as a principled alternative to such highly tuned learning rate schedules, we now present the results we obtain when training our Transformers using the following learning rate schedule:
\begin{equation}
\label{eq:noam_schedule}
	\textrm{lr}(t) \triangleq d_{\textrm{embedding}}^{-0.5} \min{\left(t^{-0.5}, t \cdot T_{\textrm{warmup}}^{-1.5}\right)},
\end{equation}
where $t$ is the current training step, $d_{\textrm{embedding}}$ is the word embeddings size, and $T_{\textrm{warmup}}$ is the number of warmup steps and is set to 10,000 in these experiments.
This schedule was proposed in the original Transformer paper \citep{Vaswani:2017}, and was tuned for the WMT dataset.

The results obtained when using this learning rate schedule are also shown in table \ref{tab:results}, under the name ``Plain*''.
{\em In both cases, our curriculum learning approach obtains a better model in about 70\% less training time}.
This is very important, especially when applying Transformers in new datasets, because such learning rate heuristics often require careful tuning.
This tuning can be both very expensive and time consuming, often resulting in very complex mathematical expressions, with no clear motivation or intuitive explanation \citep{Chen:2018}.
Our curriculum learning approach achieves better results, in significantly less time, while only requiring one parameter (the length of the curriculum). % which is not hard to set.

Note that even without using any learning rate schedule, our curriculum methods were able to achieve performance comparable to the ``Plain*'' in about twice as many training steps.
``Plain'' was not able to achieve a BLEU score above $2.00$ even after fives times as many training steps, at which point we stopped these experiments.

%\vspace{-0.5em}
\paragraph{Implementation and Reproducibility.}

We are releasing an implementation of our proposed method and experiments built on top of the machine translation library released by \citet{Platanios:2018}, using TensorFlow Scala \citep{Platanios:2018b}, and is available at {\small \url{https://github.com/eaplatanios/symphony-mt}}.
Furthermore, all experiments can be run on a machine with a single Nvidia V100 GPU, and 24 GBs of system memory.
Our most expensive experiments --- the ones using Transformers on the WMT-16 dataset --- take about 2 days to complete, which would cost about \$125 on a cloud computing service such as Google Cloud or Amazon Web Services, thus making our results reproducible, even by independent researchers.

%% file: related_work.tex
\section{Related work}
\label{sec:related-work}

The idea of teaching algorithms in a similar manner as humans, from easy concepts to more difficult ones, has existed for a long time~\citep{elman1993learning, krueger2009flexible}.
Machine learning models are typically trained using stochastic gradient descent methods, by uniformly sampling mini-batches from the pool of training examples, and using them to compute updates for the model parameters.
Deep neural networks, such as RNNs and Transformers, have highly non-convex loss functions.
This makes them prone to getting stuck in saddle points or bad local minima during training, often resulting in long training times and bad generalization performance.
\citet{Bengio:2009} propose a curriculum learning approach that aims to address these issues by changing the mini-batch sampling strategy.
They propose starting with a distribution that puts more weight on easy samples, and gradually increase the probability of more difficult samples as training progresses, eventually converging to a uniform distribution.
They demonstrate empirically that such curriculum approaches indeed help decrease training times and sometimes even improve generalization. %, for deep neural networks.

Perhaps the earliest attempt to apply curriculum learning in MT was made by \citet{Zou:2013}.
The authors employed a curriculum learning method to learn Chinese-English bilingual word embeddings, which were subsequently used in the context of phrase-based machine translation.
They split the word vocabulary in 5 separate groups based on word frequency, and learned separate word embeddings for each of these groups in parallel.
Then, they merged the 5 different learned embeddings and continued training using the full vocabulary.
While this approach makes use of some of the ideas behind curriculum learning, it does not directly follow the original definition introduced by \citet{Bengio:2009}. 
Moreover, their model required 19 days to train.
There have also been a couple of attempts to apply curriculum learning in NMT that were discussed in section~\ref{sec:intro}.

There also exists some relevant work in areas other than curriculum learning.
\citet{zhang2016boosting} propose training neural networks for NMT by focusing on hard examples, rather than easy ones.
They report improvements in BLEU score, while only using the hardest 80\% training examples in their corpus.
This approach is more similar to boosting by \citet{Schapire:1999}, rather than curriculum learning, and it does not help speed up the training process; it rather focuses on improving the performance of the trained model.
The fact that hard examples are used instead of easy ones is interesting because it is somewhat contradictory to that of curriculum learning.
Also, in contrast to curriculum learning, no ordering of the training examples is considered.

Perhaps another related area is that of active learning, where the goal is to develop methods that request for specific training examples.
\citet{Haffari:2009}, \citet{Bloodgood:2010}, and \citet{Ambati:2012} all propose methods to solicit training examples for MT systems, based on the occurrence frequency of n-grams in the training corpus.
The main idea is that if an n-gram is very rare in the training corpus, then it is difficult to learn to translate sentences in which it appears.
This is related to our sentence rarity difficulty metric and points out an interesting connection between curriculum learning and active learning.

Regarding training Transformer networks, \citet{Shazeer:2018} perform a thorough experimental evaluation of Transformers, when using different optimization configurations.
They show that a significantly higher level of performance can be reached by not using momentum during optimization, as long as a carefully chosen learning rate schedule is used.
Such learning rate schedules are often hard to tune because of the multiple seemingly arbitrary terms they often contain.
Furthermore, \citet{Popel:2018} show that, when using Transformers, increasing the batch size results in a better model at convergence.
We believe this is indicative of very noisy gradients when starting to train Transformers and that higher batch sizes help increase the signal-to-noise ratio.
We show that our proposed curriculum learning method offers a more principled and robust way to tackle this problem.
Using our approach, we are able to train Transformers to state-of-the-art performance, using small batch sizes and without the need for peculiar learning rate schedules, which are typically necessary.

%% file: conclusion.tex
\section{Conclusion and Future Work}
\label{sec:conclusion}

We have presented a novel {\em competence-based} curriculum learning approach for training neural machine translation models.
Our resulting framework is able to boost performance of existing NMT systems, while at the same time significantly reducing their training time.
It differs from previous approaches in that it does not depend on multiple hyperparameters that can be hard to tune, and it does not depend on a manually designed discretized training regime.
We define the notions of {\em competence}, for a learner, and {\em difficulty}, for the training examples, and propose a way to filter training data based on these two quantities.
Perhaps most interestingly, we show that our method makes training Transformers faster and more reliable, but has a much smaller effect in training RNNs.

In the future, we are mainly interested in: (i) exploring more difficulty heuristics, such as measures of alignment between the source and target sentences \citep{Kocmi:2017}, sentence length discrepancies, or even using a pre-trained language model to score sentences, which would act as a more robust replacement of our sentence rarity heuristic, and (ii) exploring more sophisticated competence metrics that may depend on the loss function, the loss gradient, or on the learner's performance on held-out data.
Furthermore, it would be interesting to explore applications of curriculum learning to multilingual machine translation (e.g., it may be easier to start with high-resource languages and move to low-resource ones later on).
We would also like to explore the usefulness of our framework in more general machine learning tasks, outside of NMT.